\documentclass[nohyperref]{article}

\usepackage{microtype}
\usepackage{graphicx}
\usepackage{subfigure}
\usepackage{booktabs} % for professional tables
\usepackage{multicol}

\usepackage{hyperref}

\usepackage[accepted]{icml2023}

\usepackage{amsmath}
\usepackage{amssymb}
\usepackage{mathtools}
\usepackage{amsthm}

\usepackage[capitalize,noabbrev]{cleveref}

\usepackage{color}
\usepackage{algorithm}
\usepackage[noend]{algpseudocode}
\usepackage{mathrsfs}
\usepackage{dsfont}
\usepackage{lmodern}
\usepackage{array}
\usepackage{bbm}
\usepackage{amsmath}
\usepackage{amssymb}
\usepackage[mathscr]{eucal}
\usepackage{graphicx}
\usepackage{mathrsfs}
\usepackage{psfrag}
\usepackage{color}
\usepackage{here}
\usepackage{wasysym}
\usepackage{enumitem}
\usepackage{amsthm}
\usepackage{lipsum}
\usepackage{todonotes}
\usepackage{tabulary}
\usepackage{pifont}
\usepackage{outlines}
\usepackage{thmtools, thm-restate}

\newcommand{\Prob}{\mathds{P}}
\newcommand{\Real}{\mathds{R}}

\DeclareMathOperator*{\argmax}{arg\,max}
\DeclareMathOperator*{\softmax}{softmax}

\newcommand{\Zc}{\mathcal{Z}}
\newcommand{\Ic}{\mathcal{I}}

\newcommand{\Dc}{\mathcal{D}}

\definecolor{ian_highlight}{RGB}{100, 2, 2}

\errorcontextlines\maxdimen

\makeatletter
    \newcommand*{\algrule}[1][\algorithmicindent]{\makebox[#1][l]{\hspace*{.5em}\thealgruleextra\vrule height \thealgruleheight depth \thealgruledepth}}%
\newcommand*{\thealgruleextra}{}
\newcommand*{\thealgruleheight}{.75\baselineskip}
\newcommand*{\thealgruledepth}{.25\baselineskip}

\newcount\ALG@printindent@tempcnta
\def\ALG@printindent{%
    \ifnum \theALG@nested>0% is there anything to print
        \ifx\ALG@text\ALG@x@notext% is this an end group without any text?
        \else
            \unskip
            \addvspace{-1pt}% FUDGE to make the rules line up
            \ALG@printindent@tempcnta=1
            \loop
                \algrule[\csname ALG@ind@\the\ALG@printindent@tempcnta\endcsname]%
                \advance \ALG@printindent@tempcnta 1
            \ifnum \ALG@printindent@tempcnta<\numexpr\theALG@nested+1\relax% can't do <=, so add one to RHS and use < instead
            \repeat
        \fi
    \fi
    }%
\usepackage{etoolbox}
\makeatother

\newbox\statebox
\newcommand{\myState}[1]{%
    \setbox\statebox=\vbox{#1}%
    \edef\thealgruleheight{\dimexpr \the\ht\statebox+1pt\relax}%
    \edef\thealgruledepth{\dimexpr \the\dp\statebox+1pt\relax}%
    \ifdim\thealgruleheight<.75\baselineskip
        \def\thealgruleheight{\dimexpr .75\baselineskip+1pt\relax}%
    \fi
    \ifdim\thealgruledepth<.25\baselineskip
        \def\thealgruledepth{\dimexpr .25\baselineskip+1pt\relax}%
    \fi
    \State #1%
    \def\thealgruleheight{\dimexpr .75\baselineskip+1pt\relax}%
    \def\thealgruledepth{\dimexpr .25\baselineskip+1pt\relax}%
}
\newcount\Comments
\newcommand{\kibitz}[2]{\ifnum\Comments=1{\textcolor{#1}{\textsf{\footnotesize #2}}}\fi}

\usepackage{hyperref}
\usepackage{url}

\newcommand{\githubenn}{\url{https://github.com/deepmind/enn}}

\newcommand{\githubtestbed}{\url{https://github.com/deepmind/neural\_testbed}}
\newcommand{\githubtestbedpublic}{\url{https://github.com/deepmind/neural\_testbed}}

\icmltitlerunning{Fine-Tuning Language Models via Epistemic Neural Networks}

\begin{document}

\twocolumn[
\icmltitle{Fine-Tuning Language Models via Epistemic Neural Networks}



\icmlsetsymbol{equal}{*}

\begin{icmlauthorlist}
\icmlauthor{Ian Osband}{zzz}
\icmlauthor{Mohammad Asghari}{zzz}
\icmlauthor{Benjamin Van Roy}{zzz}

\icmlauthor{Nat McAleese}{zzz}
\icmlauthor{John Aslanides}{zzz}
\icmlauthor{Geoffrey Irving}{zzz}
\end{icmlauthorlist}

% \icmlaffiliation{xxx}{DeepMind, Scalable Alignment Team, London}
% \icmlaffiliation{yyy}{DeepMind, Efficient Agent Team, Mountain View}
\icmlaffiliation{zzz}{DeepMind}

\icmlcorrespondingauthor{Ian Osband}{iosband@deepmind.com}

\icmlkeywords{Machine Learning, ICML}

\vskip 0.3in
]


\vspace{-1mm}
\printAffiliationsAndNotice{}  % leave blank if no need to mention equal contribution
\vspace{-1mm}

%%%%%%%%%%%%%%%%%%%%%%%%%%%%%%%%%%%%%%%%%%%%%%%%%%%%%%%%%%%%%%%%%%%%%%%%%%%%%%%% ABSTRACT
%%%%%%%%%%%%%%%%%%%%%%%%%%%%%%%%%%%%%%%%%%%%%%%%%%%%%%%%%%%%%%%%%%%%%%%%%%%%%%%%
\vspace{-1mm}
\begin{abstract}
% \vspace{-1mm}
% Large language models are now part of a powerful new paradigm in machine learning.
% These models learn a wide range of capabilities from training on large unsupervised text corpora.
% In many applications, these capabilities are then \textit{fine-tuned} through additional training on specialized data to improve performance in that setting.
% In this paper, we augment these models with an \textit{epinet}: a small additional network architecture that helps to estimate model uncertainty and form an \textit{epistemic neural network} (ENN).
% ENNs are neural networks that can know what they don't know.
% We show that, using an epinet to prioritize uncertain data, we can fine-tune BERT on GLUE tasks to the same performance while using 2x less data.
% We also investigate performance in synthetic neural network generative models designed to build understanding.
% In each setting, using an epinet outperforms heuristic active learning schemes.

% \vspace{-1mm}
Language models often pre-train on large unsupervised text corpora, then \textit{fine-tune} on additional task-specific data.
However, typical fine-tuning schemes do not prioritize the examples that they tune on.
We show that, if you can prioritize informative training data, you can achieve better performance while using fewer labels.
To do this we augment a language model with an \textit{epinet}: a small additional network that helps to estimate model uncertainty and forms an \textit{epistemic neural network} (ENN).
ENNs are neural networks that can know what they don't know.
Using an epinet to prioritize uncertain data, we can fine-tune BERT on GLUE tasks to the same performance while using 2x less data than training without prioritization.
We also investigate performance in synthetic neural network generative models designed to build understanding.
In each setting, using an epinet outperforms heuristic active learning schemes.
% \vspace{-1mm}
\end{abstract}
% \vspace{-2mm}

%%%%%%%%%%%%%%%%%%%%%%%%%%%%%%%%%%%%%%%%%%%%%%%%%%%%%%%%%%%%%%%%%%%%%%%%%%%%%%%% INTRODUCTION
%%%%%%%%%%%%%%%%%%%%%%%%%%%%%%%%%%%%%%%%%%%%%%%%%%%%%%%%%%%%%%%%%%%%%%%%%%%%%%%%
\vspace{-4mm}
\section{Introduction}
\label{sec:intro}
% \vspace{-1mm}

% \begin{outline}
% \1 Set the scene for why we are interested in fine-tuning language models.
%     \2 Mention LLM \citep{brown2020language}.
%     \2 Mention challenges in algnment \citep{ziegler2019fine, ouyang2022training}.
%     \2 High level AGI dreams.
%     \fillpara
Large language models (LLMs) have emerged as one of the most exciting new developments in artificial intelligence research \citep{brown2020language}.
Typically, large autoregressive transformers \citep{vaswani2017attention} are trained on large corpora of internet text data simply to predict the next token in an unsupervised manner.
In spite of this simple training objective, these models have gone on to demonstrate impressive performance across a wide variety of tasks, and under different evaluation protocols \citep{rae2021scaling}.
Importantly, these capabilities seem to improve with both increased computation and increased training data \citep{kaplan2020scaling, hoffmann2022training}.

% \1 Fine-tuning, alignment, active learning.
%     \2 We need to know what we don't know.
%     \2 Can use this for safety, decision making, and active learning.
%     \fillpara
LLMs are famously able to demonstrate impressive few-shot learning capabilities, induced by their ability to complete specially-designed text `prompts' \citep{brown2020language}.
However, these capabilities can typically be improved by \textit{fine-tuning} the model on additional training data specialized to the particular application of interest \citep{ziegler2019fine, cobbe2021training}.
Beyond raw capability, this fine-tuning is also essential for language model \textit{alignment}: to make sure the system accurately reflects the goals of the algorithm designer \citep{ouyang2022training}.
This setting has even been proposed as an important case-study for the more general challenge of AI alignment and existential risk \citep{askell2021general}.

% \1 Uncertainty estimates in large language models.
%     \2 Previous investigations suggested it was not possible \citep{gleave2022uncertainty}.
%     \2 Epistemic neural networks \citep{osband2021epistemic}.
%     \2 The idea is that we can get better uncertainty estimates, and then use them to be better.
%     \fillpara
One of the key challenges at the heart of the alignment problem comes from dealing with \textit{uncertainty}.
LLMs, as they are currently instantiated, are conventional neural networks that do not distinguish irreducible uncertainty over the next token (e.g. when the outcomes is produced by a rolling a die), from uncertainty that could be resolved with more training data (e.g. when the outcome is determined by recent history but the model has not been trained enough to know this).
This distinction is sometimes referred to as the distinction between \textit{aleatoric} (relating to chance) and \textit{epistemic} (relating to knowledge) uncertainty \citep{KendallGal17WhatUncertainties}.
Even though experimental results suggest LLMs can be trained to estimate uncertainty over the validity of their claims \citep{kadavath2022language}, this does not address this key distinction in the \textit{sources} of uncertainty in ways that are necessary to accelerate learning.
Previous attempts to augment LLMs with ensemble-based uncertainty have not been successful, and suggest that maybe a new training paradigm is necessary to incorporate this distinction \citep{gleave2022uncertainty}.

\begin{figure}[!ht]
    \vspace{-2mm}
    \centering
    \includegraphics[width=0.99\columnwidth{}]{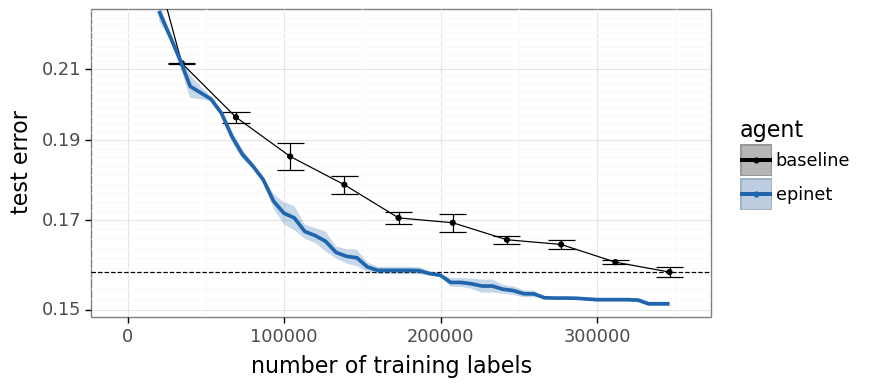}
    \vspace{-5mm}
    \caption{Fine-tuning a BERT model on the MNLI dataset \citep{williams2017broad}. The baseline BERT is tuned over learning rate, batch size and number of SGD steps for each random subset of training labels. The baseline does not prioritize training data.  Using an epinet to prioritze, we are able to match baseline performance on the entire training set (the dashed line) with 2x less data (Section~\ref{sec:language}).}
    \label{fig:mnli_headline}
    \vspace{-2mm}
\end{figure}

% \1 We show that this actually works: better uncertainty can lead to better fine-tuning.
%     \2 Emphasize importance of good \textit{joint} predictions \citep{wen2022predictions}.
%     \2 Show on a progression from theory, to synthetic data, to language models.
%     \2 Figure \ref{fig:mnli_headline} shows results are impressive, and we get 2x reduction in data requirements.
%     \fillpara
In this paper, we explore an alternative approach to supplement LLMs with an \textit{epinet}: a small additional network trained to estimate model uncertainty \citep{osband2021epistemic}.
Taken together, the base LLM and epinet form an \textit{epistemic neural network} (ENN).
ENNs are neural networks that can know what they don't know.
Figure~\ref{fig:mnli_headline} offers a preview of the results of Section~\ref{sec:language}, where we examine a model based on BERT fine-tuned on GLUE datasets \citep{devlin2018bert, wang2018glue}.
We consider a baseline that does not prioritize training data.  \textbf{We see that, using an epinet to prioritize training labels, we can match the performance of this baseline with 2x less data.}
We also see significant improvements in final accuracy from using an an epinet.

%%%%%%%%%%%%%%%%%%%%%%%%%%%%%%%%%%%%%%%%%%%%%%%%%%%%%%%%%%%%%%%%%%% Key contributions
\subsection{Key contributions}
\label{sec:intro_key}
\vspace{-1mm}

% \begin{outline}

% \1 We show that effective uncertainty estimates can drive active learning in language models.
%     \2 We provide a consistent theory, with development building up from toy to BERT.
%     \fillpara
We show that \textbf{LLMs can be augmented with an epinet to deliver effective results in a computationally scalable manner}.
We first present results with a synthetic neural-net generative model \cite{osband2022neural}.  This model has been used previously to demonstrate that the epinet produces useful estimates of epistemic uncertainty \cite{osband2021epistemic}.  We show that prioritizing data using an epinet-enhanced model reduces the number of training labels required to match baseline performance.  We then demonstrate a similar outcome fine-tuning BERT, where we obtain baseline performance using 2x less data.

% \1 We show that prioritizing by uncertainty can even be better than many approaches that rely on the true label.
%     \2 Comparing against loss, gradient of loss... which don't seem to work very well!
%     \2 Margin/entropy methods do better, but we also outperform those.
%     \fillpara
We demonstrate \textbf{improvement over prioritizing based on other heuristic approaches}.
Prior work has suggested that prioritizing by LLM class probabilities, which do not distinguish epistemic from aleatoric uncertainty, can be an effective heuristic for active learning.
We reproduce these results, but find that our epinet-enhanced model leads to further improvement.
    
% \1 We show that the novel ENN epinet can be more effective than popular Bayesian deep learning approaches.
%     \2 People said you couldn't do ensemble... but you actually can.
%     \2 Epinet is much better than ensemble and dropout from a single base model.
%     \fillpara
% \end{outline}
Finally, \textbf{the epinet is at least as effective as existing approaches to Bayesian deep learning}.
Where prior work had concluded that ensemble approaches to uncertainty estimation were not effective for LLMs, we find that both ensemble-based approaches and dropout can also drive improved active learning \citep{gleave2022uncertainty}.
However, even when compared to these well-tuned baselines, we find that using an epinet is able to match performance at a lower computational cost.
Although this paper focuses on \textit{data} efficiency, computational concerns may be an important consideration in other applications.

%%%%%%%%%%%%%%%%%%%%%%%%%%%%%%%%%%%%%%%%%%%%%%%%%%%%%%%%%%ayesiand%%%%%%%%% Related work
\vspace{-1mm}
\subsection{Related work}
\label{sec:intro_related}
\vspace{-1mm}

Our work is motivated by the emerging importance of large pre-trained models across language \citep{brown2020language}, images \citep{radford2021learning} and beyond \citep{bommasani2021opportunities}.
While these `foundation models' are trained on huge amount of unsupervised data, their performance on downstream tasks can be typically boosted by fine-tuning on additional task-specific examples \citep{ziegler2019fine, cobbe2021training}.
Fine-tuning a `foundation model' can be much more sample efficient than re-training from scratch, but may still require substantial amounts of data \citep{Stiennon2020learning}.

% \1 Uncertainty and deep learning.
% \fillpara
The field of \textit{active learning} has studied the problem of optimizing which data to collect \citep{settles2009active}.
Research in this area has unearthed a wide variety of heuristic approaches \citep{lewis1995sequential, seung1992query}, and which can be successful in particular benchmarks \citep{settles2008analysis, sadigh2017active, beluch2018power}.
The problem has been most comprehensively studied through a Bayesian lens, through which the problem can naturally be framed as one to maximize the \textit{information gain} through training data \citep{houlsby2011bayesian}.
While the Bayesian perspective offers a coherent statistical perspective, the computational costs of exact Bayesian inference can become intractable even for small systems \citep{welling2011bayesian}.

% \1 Fine tuning and aligment.
% \fillpara
Our paper seeks to find tractable computational solutions to active learning that are compatible with LLMs.
Related work has studied similar approximation schemes applied to deep learning, but with a focus on image datasets \citep{gal2017deep, christiano2017deep, kirsch2019batchbald}.
However, other papers have shown that the proposed method of dropout for Bayesian inference can be of very poor quality \citep{osband2016risk, hron2017variational, osband2022neural}.
These results have been complemented by an empirical evaluation that concluded a fundamental rethink was necessary for effective uncertainty in LLMs \citep{gleave2022uncertainty}.
In this paper, we build on the recent development of \textit{epistemic neural networks} (ENNs), which can offer a new and more effective approach for neural networks that know what they don't know \citep{osband2021epistemic}.
Our work complements other recent investigations into the feasibility of active learning with LLMs \citep{dor2020active,margatina2022importance}.
Both of these papers show that heuristic methods for active learning can improve data efficiency in fine-tuning.
Our work reproduces this finding, and shows that prioritization based on more principled uncertainty estimates, as afforded by ENNs, can be even more effective.

% \end{outline}

%%%%%%%%%%%%%%%%%%%%%%%%%%%%%%%%%%%%%%%%%%%%%%%%%%%%%%%%%%%%%%%%%%%%%%%%%%%%%%%% PROBLEM FORMULATION
%%%%%%%%%%%%%%%%%%%%%%%%%%%%%%%%%%%%%%%%%%%%%%%%%%%%%%%%%%%%%%%%%%%%%%%%%%%%%%%%
\section{Problem formulation}
\label{sec:problem}
% \vspace{-1mm}

This section reviews the notation necessary to describe the agents we consider.
We formalize the active learning problem, and review benchmark approaches to label prioritization in the literature.
As part of our submission we also open-source our code in Appendix~\ref{app:code}.

% \vspace{-1mm}
%%%%%%%%%%%%%%%%%%%%%%%%%%%%%%%%%%%%%%%%%%%%%%%%%%%%%%%%%%%%%%%%%%% Epistemic neural networks
\subsection{Epistemic neural networks}
\label{sec:problem_epistemic}
% \vspace{-1mm}

% \1 Outline the framing of epistemic neural networks \citep{osband2021epistemic}.
%     \2 Set up the minimal notation so that we can explain our algorithms.
%     \fillpara
We examine the performance of agents based on epistemic neural networks (ENN) \citep{osband2021epistemic}.
A conventional neural network is specified by a parameterized function class $f$, which produces an output $f_\theta(x)$ given parameters $\theta$ and an input $x$.
An ENN is specified by a parameterized function class $f$ \textit{and} a reference distribution $P_Z$.
The output $f_\theta(x, z)$ of an ENN depends additionally on an \textit{epistemic index} $z$, sampled from the reference distribution $P_Z$.
Variation of the network output with $z$ indicates uncertainty that might be resolved by future data.
All conventional neural networks can be written as ENNs, but this more general framing allows an ENN to represent the kinds of uncertainty necessary for useful joint predictions \citep{wen2022predictions}.

% \1 This gives you a way of making joint predictions, but not necessarily Bayesian.
ENNs share many similarities with Bayesian neural networks (BNNs), which maintain a posterior distribution over plausible neural nets.
However, unlike BNNs, ENNs do not necessarily ascribe Bayesian semantics to the unknown parameters of interest, and do not generally update with Bayes rule.
All BNNs can be expressed as ENNs;
for example, an ensemble of $K$ networks $f_{\theta_1}, .., f_{\theta_K}$ can be written as an ENN $\tilde{f}$ with index $z \sim {\rm Unif}(\{1,..,K\})$ and $\tilde{f}_\theta(x, z) := f_{\theta_z}(x)$ \citep{osband2015bootstrapped, lakshminarayanan2017simple}.
However, there are some ENNs that cannot be expressed naturally as BNNs.

% \1 This gives you a way of making joint predictions, but not necessarily Bayesian.
One such example of novel ENNs is the \textit{epinet}: a small additional network designed to estimate uncertainty.
An epinet is added to a \textit{base network}: a conventional NN with base parameters $\zeta$ and input $x$ with output $\mu_\zeta(x)$.
The epinet acts on a subset of \textit{features} $\phi_\zeta(x)$ derived from the base network, as well as an epistemic index $z$ sampled from the standard normal in $D_Z$ dimensions.
For concreteness, you might think of $\mu_\zeta(\cdot)$ as an LLM, with features $\phi_\zeta(x)$ as the final hidden layer in the model.
For epinet parameters $\eta$, this produces a combined output: \vspace{-2mm}
\begin{equation}
\label{eq:epinet}
\underbrace{f_\theta(x, z)}_{\text{ENN}} = \underbrace{\mu_\zeta(x)}_{\text{base net}} + \underbrace{\sigma_\eta(\mathrm{sg}[\phi_\zeta(x)], z)}_{\text{epinet}}.
\end{equation}
The ENN parameters $\theta = (\zeta, \eta)$ include those of the base network and epinet.
The epinet $\sigma_\eta$ is a simple MLP architecture, with an internal \textit{prior function} designed to create an initial variation in index $z$ \citep{osband2018rpf}.
The ``stop gradient'' notation $\mathrm{sg}[\cdot]$ indicates the argument is treated as fixed when computing a gradient. For example, 
$\nabla_\theta f_\theta(x,z) = \left[\nabla_\zeta \mu_\zeta(x), \nabla_\eta \sigma_\eta(\phi_\zeta(x), z) \right]$.
This approach has been shown to outperform ensembles of hundreds of particles at a computational cost less than that of two particles.
For more detail on the epinet we recommend reviewing \cite{osband2021epistemic}.

%%%%%%%%%%%%%%%%%%%%%%%%%%%%%%%%%%%%%%%%%%%%%%%%%%%%%%%%%%%%%%%%%%% Active learning
\subsection{Active learning}
\label{sec:problem_active}
\vspace{-1mm}

% Overview of the active learning problem
We consider the active learning setting, where a learning agent is able to prioritize training examples in order to improve performance on held out data.
More formally, we consider a training dataset $\Dc = \{(x_i, y_i, i)\}_{i=1}^N$ for inputs $x_i$ and class labels $y_i$.
Initially, at timestep $t=0$, the agent can observe the training inputs $\Dc_X = \{(x_i, i)\}_{i=1}^N$ but none of the associated class labels.
At each timestep $t$, the agent selects $a_t \in \{1,..,N\}$ and reveals the corresponding class label $y_{a_t}$.
The agent can use the data $\Dc_{t-1} = \Dc_X \cup \{y_{a_s}\}_{s=0}^{t-1}$ to update its beliefs, which we encode in the ENN parameters $\theta_t$.

% How to assess performance in a learning problem
We assess the quality of active learning agents through their loss evaluated on a held-out dataset  $\tilde{\Dc} = \{(\tilde{x}_i, \tilde{y}_i, i) \}_{i=1}^{\tilde{N}}$.
Examples of such evaluation might include classification error or log-loss evaluated on a test set.
Agents that successfully prioritize their training labels to achieve lower held-out loss with fewer training labels can be said to perform better.
Our next subsection provides some concrete approaches to prioritizing training labels.

%%%%%%%%%%%%%%%%%%%%%%%%%%%%%%%%%%%%%%%%%%%%%%%%%%%%%%%%%%%%%%%%%%% Label selection
\subsection{Priority functions}
\label{sec:problem_priority}
\vspace{-1mm}

% \begin{outline}
% \1 This is going to outline some of the ways in which we can select data.
%     \2 Mainly we're just trying to be able to explain what we run in the experiments.
%     \fillpara

We focus on active learning schemes that prioritize training labels based on a \textit{priority function} $g$.
The priority function maps ENN parameters $\theta$ and input $x$ to a score $g(\theta, x) \in \Real$.
We focus our attention on the problem of classification, where the ENN makes predictions over class labels $c \in \{1, .., C\}$.
We introduce notation for the class probabilities as predicted by an ENN, \vspace{-2mm}
\begin{eqnarray}
    p(c | \theta, x, z) := \softmax(f_{\theta}(x, z))_c, \\
    p(c | \theta, x) := \int_z P_Z(dz) p(c | \theta, x, z).
\end{eqnarray}
For any probability distribution $p$ over $C$ classes we recall that the \textit{entropy} is defined
\mbox{$\mathbb{H}[p] := -\sum_{c} p(c) \log p(c)$} \citep{shannon2001mathematical}.
We can now describe the priority functions used in this paper.
The key baseline for our experiment comes from the \textbf{uniform} prioritization, which does not distinguish between inputs,
%which does not distinguish based on input,
$g^{\rm uniform} = 0$.
% \begin{equation}
%     g^{\rm uniform}(\theta, x) = 0.
% \end{equation}

% \begin{itemize}[leftmargin=*]

% \item \textbf{uniform}: this baseline method assigns score uniformly at random

% \end{itemize}

Our next two priorities require only conventional neural network predictions, and do not account for variability in the epistemic index $z$.
We call these approaches \textit{marginal} priority functions since they only depend on the marginal probability estimates for a single input $x$.
We define the \textbf{entropy} prioritization $g^{\rm entropy}(\theta, x) = \mathbb{H}[p(\cdot | x, \theta)]$
% \begin{equation}
%     g^{\rm entropy}(\theta, x) = \mathbb{H}[p(\cdot | x, \theta)].
% \end{equation}
Similarly,  \textbf{margin} \citep{roth2006margin} prioritizes based on the smallest difference between the highest and second highest class probabilities,
$g^{\rm margin}(\theta, x) = p (c_2 | \theta, x) - p (c_1 | \theta, x),$
% \begin{equation}
% \label{eq:margin}
%     g^{\rm margin}(\theta, x) = p (c_2 | \theta, x) - p (c_1 | \theta, x),
% \end{equation}
where $c_1 \in \argmax_{c} p (c | \theta, x)$ and $c_2 \in \argmax_{c \neq c_1} p(c | \theta, x)$.
Both of these priorities are maximized by the uniform distribution, but neither distinguish genuinely ambiguous examples, from those with insufficient data \citep{osband2021epistemic}.

Our final approaches use the ENN to prioritize uncertain, as opposed to ambiguous, datapoints.
We can call these priority functions \textit{epistemic} priority functions, since they depend on the joint distribution of predictions.
We define \textbf{bald} \citep{houlsby2011bayesian} priority based on the mutual information gain:
$ g^{\rm bald}(\theta, x) = \mathbb{H}[p(\cdot | \theta, x)]  - \int_z P_Z(dz) \mathbb{H}[ p(\cdot | \theta, x, z)],$
% \begin{equation}
% \label{eq:bald}
% g^{\rm bald}(\theta, x) = \mathbb{H}[p(\cdot | \theta, x)] 
% - \int_z P_Z(dz) \mathbb{H}[ p(\cdot | \theta, x, z)],
% \end{equation}
Similarly, \textbf{variance} uses the variation in predicted probabilities,
$g^{\rm variance}(\theta, x) = \sum_{c}  \int_z P_Z(dz) \left( p(c | \theta, x, z) - p(c | \theta ,x) \right)^2.$
% \begin{equation}
% \label{eq:variance}
% g^{\rm variance}(\theta, x) = \sum_{c}  \int_z P_Z(dz) \left( p(c | \theta, x, z) - p(c | \theta ,x) \right)^2.
% \end{equation}
Both of these methods prefer to select training examples with high variability in prediction in the epistemic index $z$.

%%%%%%%%%%%%%%%%%%%%%%%%%%%%%%%%%%%%%%%%%%%%%%%%%%%%%%%%%%%%%%%%%%% Label selection
\subsection{Training algorithm}
\label{sec:problem_training}

% Explain the slightly strange version of active learning that we do
In our experiments, we consider a variant of stochastic gradient descent that proceeds in steps $s=0,1,2,..$.
At gradient step $s$, a large candidate batch of size $N_B$ is selected uniformly at random from the training dataset $\Dc$ without replacement.
Then, the agent selects $N_b \le N_B$ elements of this candidate batch to perform SGD upon.
Depending on the choice of data selected, this can require between $0$ and $N_b$ new training labels to be revealed.
It is most typical in the active learning literature to consider the case $N_B=N$ and $N_b=1$.
However, in large scale distributed frameworks it can be useful to consider other variants that don't require passing through the entire dataset $N_B < N$, and which may select more than one training datapoint at a time $N_b > 1$ \citep{kirsch2019batchbald}.

Algorithm~\ref{alg:enn_training} formalizes the training procedure we use for the experiments in this paper.
For all of the experiments in this paper we use a standard cross-entropy loss term with L2 regularization,
{
\medmuskip=0mu
\thinmuskip=0mu
\thickmuskip=0mu
\begin{equation}
\label{eq:xent}
    \ell_\lambda^{\rm XENT}(\theta, z, x_i, y_i, i) := - \ln\left(\softmax(f_\theta(x_i, z))_{y_i}\right) + \lambda\|\theta\|_2^2.
\end{equation}
}
\hspace{-1mm}Here, $\lambda$ is a hyperparameter that scales the regularization penalty.
% As demonstrated by \cite{dwaracherla2022ensembles}, it can sometimes be beneficial to randomly perturb the loss function via versions of the statistical bootstrap \citep{efron1994introduction}.
% We were able to obtain good results in our problems without bootstrapping, and leave that open for future work.
Algorithm~\ref{alg:enn_training} is not meant to be the best possible active learning scheme, but provides a lower bound on what is possible with active learning.
We compare this approach varying the priority function and ENN like-for-like, as well as a non-prioritized baseline together with optimal early stopping.
We see that, compared to both baselines, the epinet-enabled model is able to learn much faster.

\begin{figure*}[t]
\centering
\begin{minipage}[htp]{0.8\textwidth}
\begin{algorithm}[H]
\caption{ENN active learning via stochastic gradient descent}
\label{alg:enn_training}

{\small
\begin{tabular}{lll}
\textbf{Inputs:} 
& dataset & training examples $\Dc = \{(x_i, y_i, i) \}_{i=1}^N$, batch size $N_B$\\
& ENN & architecture $f$, reference distribution $P_Z$, parameters $\theta_0$\\
& priority function & $g$ evaluates parameters $\theta$ and input $x$ \\
& loss function  & $\ell$ evaluates parameters $\theta$ and index $z$ on example $(x_i, y_i, i)$\\
& batch size & learning batch size $n_b$, number of index samples $n_Z$ \\
& optimization & update rule \texttt{optimizer} and number of SGD steps $S$\\
\textbf{Returns:} & $\theta_S$ & parameter estimates for the trained ENN.
\end{tabular}

\begin{algorithmic}[1]
\For{$s$ in $0,\ldots,S-1$}
\State sample $N_B$ candidate indices $\tilde{\Ic}^C = i^C_1, .., i^C_{N_B}$ from $\{1,..,N\}$ without replacement.
\State select the $N_b$ indices with highest priority $g(\theta, x_i)$, call this $\tilde{\Ic} = i_1,..,i_{N_b} \subset \tilde{\Ic}^C$.
\State sample $n_Z$ epistemic indices $\tilde{\Zc} = z_1, .., z_{n_Z} \sim P_Z$.
\State compute minibatch $\mathtt{gradient} \leftarrow \nabla_{\theta | \theta=\theta_s} \sum_{z \in \tilde{\Zc}} \sum_{i \in \tilde{\Ic}} \ell(\theta, z, x_i, y_i, i)$.
\State update $\theta_{s+1} \leftarrow \mathtt{optimizer}(\theta_s, \mathtt{gradient})$
\EndFor
\end{algorithmic}
}
\end{algorithm}
\end{minipage}
\end{figure*}

%%%%%%%%%%%%%%%%%%%%%%%%%%%%%%%%%%%%%%%%%%%%%%%%%%%%%%%%%%%%%%%%%%%%%%%%%%%%%%%% SYNTHETIC DATA
%%%%%%%%%%%%%%%%%%%%%%%%%%%%%%%%%%%%%%%%%%%%%%%%%%%%%%%%%%%%%%%%%%%%%%%%%%%%%%%%
\section{Synthetic data}
\label{sec:synthetic}

This section evaluates active learning schemes in a simple synthetic problem designed around a neural network generative model.\footnote{
By evaluating algorithms with a controlled generative model, we can avoid some of the pitfalls of evaluation on a fixed benchmark dataset.
Benchmarks that rank performance on a fixed dataset are vulnerable to overfitting through iterative hill-climbing \citep{russo2015much}.
As such, optimized solutions on fixed datasets often don't generalize beyond the benchmark \citep{recht2018cifar}.}
We make minor modifications to the Neural Testbed \citep{osband2022neural} to create a new open-source active learning benchmark (Appendix~\ref{app:code}).
% We then evaluate different approaches to active learning in this setting.
We find that, using an epinet, we can greatly reduce the necessary amount of training labels.

%%%%%%%%%%%%%%%%%%%%%%%%%%%%%%%%%%%%%%%%%%%%%%%%%%%%%%%%%%%%%%%%%%% Environment
\subsection{Environment}
\label{sec:synthetic_environment}

% \begin{outline}
% \1 Describe the generative model / testbed set up.
% \fillpara
We consider a generative model based on data generated by a random MLP.
The inputs $x_t \in \Real^D$ are sampled from a standard normal $N(0, I)$.
Class labels are then assigned,
\begin{equation}
    \label{eq:probs}
    \Prob(y_t = y | \theta^*) = \softmax \left( h_{\theta^*}(x_t) / \rho \right)_y.
\end{equation}
Here the function $h_{\theta^*}(\cdot) \in \Real^2$ is a 2-layer MLP with ReLU activations and 50 hidden units at each layer.
The network weights $\theta^*$ are sampled according to Glorot initialization \citep{glorot2010understanding}, and $\rho > 0$ is a temperature parameter.

% \1 Specifics of kernel choice, seeds etc.
% \fillpara
In our experiments, we set input dimension $D=10$, temperature $\rho=0.1$, and generate $N=200$ training examples.
We evaluate performance through the average test log-likelihood over $\tilde{N}=1000$ testing examples, and average our results over 10 random seeds.

% \1 Some motivation for why we do this.
%     \2 Point is to have a clear and simple proof of concept, so we can build understanding
%     \fillpara
This problem is clearly a small-scale and extremely simplified model of active learning for language models.
However, we believe it is useful to begin our experimentation with a clear and simple proof of concept, so that we can build understanding in research.
While demonstrating successful active learning in toy problems does not guarantee efficacy at large scale, methods that are \textit{unable} to succeed in this setting may exhibit fundamental flaws that we should be aware of.
% In addition, by evaluating our methods on a generative model, we can avoid some of the dangers of overfitting our methods and research to any fixed dataset \citep{russo2015much,osband2022neural, osband2022evaluating}.

% \end{outline}

%%%%%%%%%%%%%%%%%%%%%%%%%%%%%%%%%%%%%%%%%%%%%%%%%%%%%%%%%%%%%%%%%%% Agent
\subsection{Agents}
\label{sec:synthetic_agent}
\vspace{-1mm}

We consider baseline agents that learn based on ENN variants of a 2-layered MLP with 50 hidden units in each layer.
We use open-source implementations in the Neural Testbed \citep{osband2022neural}.
These learning agents are tuned over multiple hyperparameters to optimize performance in this generative model, which we outline in Table~\ref{tab:agent_summary}.

\begin{table*}[!ht]
\caption{Summary of benchmark agent architectures, full details in Appendix \ref{app:synthetic}.}
\label{tab:agent_summary}
\begin{center}
\resizebox{0.8\textwidth}{!}{%
\begin{tabular}{|l|l|l|}
\hline
% \rowcolor{tableHeader}
\textcolor{black}{\textbf{agent}}          & \textcolor{black}{\textbf{description}}            & \textcolor{black}{\textbf{hyperparameters}} \\[0.5ex]  \hline
\textbf{\texttt{mlp}}            & vanilla MLP        &  $L_2$ decay                        \\
\textbf{\texttt{ensemble}}       & deep ensembles \citep{lakshminarayanan2017simple}          & $L_2$ decay, ensemble size                        \\
\textbf{\texttt{dropout}}    & dropout \citep{Gal2016Dropout}             &          $L_2$ decay, network, dropout rate                \\
% \textbf{\texttt{ensemble+}} & ensemble + prior functions  \citep{osband2018rpf}  &      $L_2$ decay, ensemble size, prior scale, bootstrap                    \\
\textbf{\texttt{epinet}}         & MLP + MLP epinet \citep{osband2021epistemic}                   &              $L_2$ decay, network, prior, index dimension \\ \hline
\end{tabular}%
}
\end{center}
\vspace{-3mm}
\end{table*}

Our results will investigate the performance of these learning agents when paired with the priority functions of Section~\ref{sec:problem_priority}.
We perform active learning according to Algorithm~\ref{alg:enn_training} by selecting $N_b=1$ data point to train on after examining all $N_B=N$ candidate training examples.
For all algorithms we use the Adam optimizer with learning rate 1e-3 \citep{kingma2014adam}.
We include links to the open-source implementations in Appendix~\ref{app:code}, together with some discussion of the technical details in Appendiex~\ref{app:synthetic}.

We compare our active learning agents against a \textbf{baseline} agent that does not perform active learning, and is not trained online.
The \textbf{baseline} agent is trained by standard supervised learning on a fraction $\psi \in \{0.01, 0.03, 0.1, 0.2, 0.3, 0.4, 0.5, 0.6, 0.7, 0.8, 0.9, 1.\}$ of the available training data.
For each of these sub-datasets, we sweep over batch size $\in \{4, 16, 64\}$, learning rate $\in \{$1e-6, 3e-6, 1e-5, 3e-5, 1e-4, 3e-4$\}$, $L_2$ regularization $\in \{$0, 1e-3, 1e-2, 1e-1, 1$\}$ and select the best SGD step over 10 training epochs in hindsight.
For each setting, we average the results over three random seeds to obtain standard error estimates for the quality of this baseline.
We then pick the best hyperparameters per data fraction.
As such, although we call this agent a \textbf{baseline}, this procedure is meant to provide an upper bound on the performance for any agent that does not prioritize its training data.

%%%%%%%%%%%%%%%%%%%%%%%%%%%%%%%%%%%%%%%%%%%%%%%%%%%%%%%%%%%%%%%%%%% Results
\subsection{Results}
\label{sec:synthetic_results}
\vspace{-1mm}
% \begin{outline}

\begin{figure}[!ht]
    \vspace{-2mm}
    \centering
    \includegraphics[width=0.95\columnwidth]{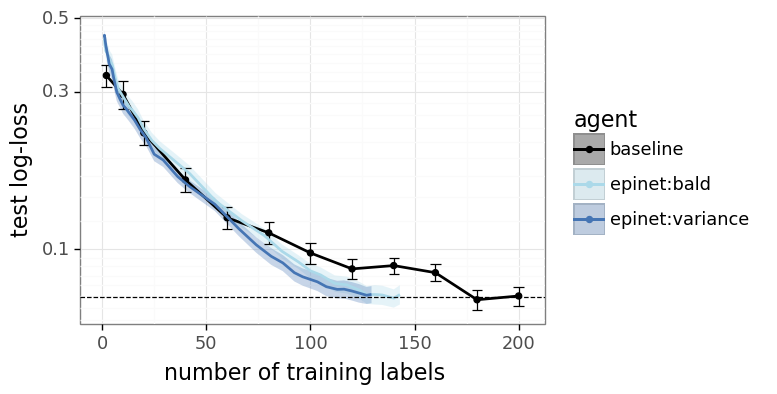}
    \vspace{-2mm}
    \caption{Learning with an epinet is able to match the baseline with fewer training labels.}
    \label{fig:testbed_headline}
    \vspace{-2mm}
\end{figure}

% \1 Describe the main agents/algorithms we have fun.
% \fillpara
Figure~\ref{fig:testbed_headline} compares learning with an epinet against our tuned supervised \textbf{baseline}.
The agents \textbf{epinet:bald} and \textbf{epinet:variance} refer to epinet trained with bald and variance priorities.
We can see that these two methods are essentially statistically indistinguishable in this setting.
Our results clearly show that, by using an epinet, the learning agent can obtain the same test performance with significantly less data.

\begin{figure}[!ht]
\centering
\subfigure[Epinet outperforms marginal methods.]
{
  \centering
  \includegraphics[width=.9\columnwidth]{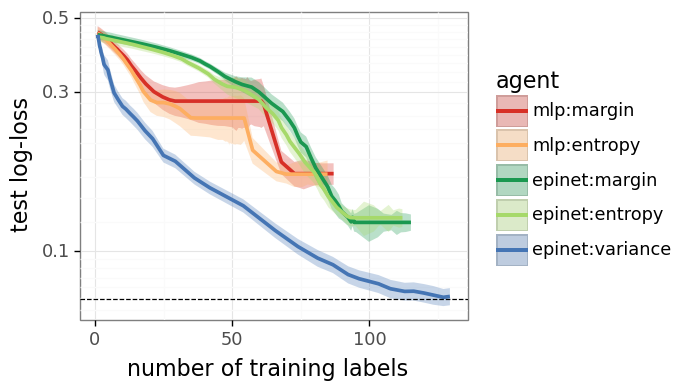}
  \label{fig:testbed_marginal}
  }
\vspace{1mm}
\subfigure[Epinet outperforms other ENNs.]
{
  \centering
  \includegraphics[width=.9\columnwidth]{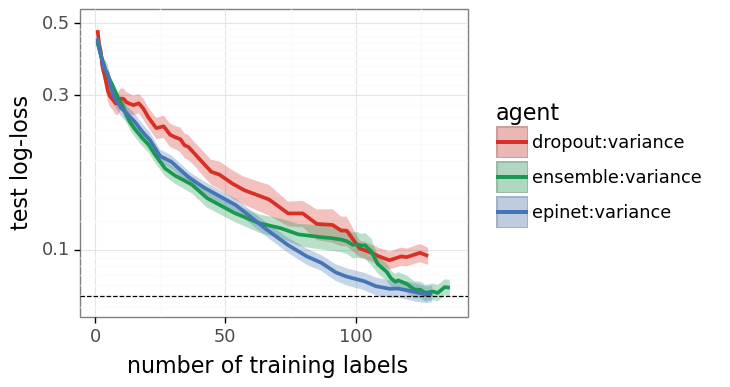}
  \label{fig:testbed_bayes}
  }
\vspace{-5mm}
\caption{Active learning with an epinet is able to provide larger improvements in learning speed than competing approaches on the Neural Testbed \citep{osband2022neural}.}
\vspace{-1mm}
\label{fig:testbed_comparison}
\end{figure}

Figure~\ref{fig:testbed_marginal} compares epinet prioritized by variance against other methods that do not make use of epistemic uncertainty.
For both MLP and epinet architectures, the choice of \textbf{margin} vs \textbf{entropy} prioritization makes little difference.
For both network architectures these methods perform significantly worse than methods that prioritize by epistemic uncertainty.
To get some intuition for how this can happen, these marginal approaches prioritize datapoints with high aleatoric uncertainty.
By contrast, agents that prioritize by epistemic uncertainty (including \textbf{bald} and \textbf{variance}) are not drawn to these points.

% \1 Another major point is that epinet has better epistemic uncertainty and so has better performance.
% \fillpara
The performance of epinet is also impressive when we compare against other approximate Bayesian approaches to uncertainty quantification.
Figure~\ref{fig:testbed_bayes} compares these agents varying only the choice of ENN architecture from \textbf{epinet} to \textbf{dropout} to \textbf{ensemble}.
We see the epinet performs better than the competing approaches, and that the ensemble performs better than dropout in this setting.
The computational cost of the epinet is much smaller than either an ensemble or dropout, since only the epinet, not the base network, needs to make a forward pass per index $z$.
These results show that in this simple and sanitized setting, active learning with an epinet can be competitive with existing approaches to Bayesian deep learning.

\vspace{-1mm}

% However, our focus in this paper is on extending these insights to learning with LLMs.
% In the next section, we will show that this same core insight carries over to more representative problems in language modeling.

%%%%%%%%%%%%%%%%%%%%%%%%%%%%%%%%%%%%%%%%%%%%%%%%%%%%%%%%%%%%%%%%%%%%%%%%%%%%%%%% LANGUAGE MODELS
%%%%%%%%%%%%%%%%%%%%%%%%%%%%%%%%%%%%%%%%%%%%%%%%%%%%%%%%%%%%%%%%%%%%%%%%%%%%%%%%
\vspace{-2mm}
\section{Language models}
\label{sec:language}
\vspace{-1mm}

In this section, we present the main empirical results of this paper.
We show that the key insights afforded by the toy problem of Section~\ref{sec:synthetic} also carry over to fine-tuning language models.
We review the problem formulation, and the experimental setup we use based on fine-tuning BERT \citep{devlin2018bert} in Jaxline \citep{deepmind2020jax}.
Using an epinet, we can greatly reduce the necessary amount of training labels.

%%%%%%%%%%%%%%%%%%%%%%%%%%%%%%%%%%%%%%%%%%%%%%%%%%%%%%%%%%%%%%%%%%% Environment
\vspace{-1mm}
\subsection{Environment}
\label{sec:language_environment}
\vspace{-1mm}

% \1 Describe the problem: BERT finetuning on GLUE tasks.
% \fillpara
The General Language Understanding Evaluation (GLUE) benchmark \citep{wang2018glue} is a collection of diverse natural language understanding tasks.
These tasks are widely accepted as a benchmark for performance in the field, and improving performance on GLUE was one of the key markers that made BERT \citep{devlin2018bert} a seminal contribution to the field.
For simplicity in a short paper, we only consider the \textit{classification} tasks of GLUE, and so exclude 3 of the 11 benchmark tasks.
We do not believe that including these tasks would pose any fundamental difficulty, but would require considering an alternative to the cross-entropy loss \eqref{eq:xent}, such as mean-squared error.

% \1 Some motivation for why we do this.
%     \2 This is not the biggest language model, but it is a classic and clear opensource.
%     \fillpara
BERT, with just over 100M parameters, is actually quite small compared to later LLMs that have scaled up to 1T+ \citep{chowdhery2022palm, fedus2021switch}.
Larger models, trained with more data, have been shown to perform better on downstream tasks \citep{kaplan2020scaling, rae2021scaling}.
These improvements have been so significant that there is a newer SuperGLUE benchmark \citep{wang2019superglue}, which replaces some of the easier tasks in GLUE \citep{wang2018glue}.
Nonetheless, we focus on BERT and GLUE because it is a clear benchmark transformer \citep{vaswani2017attention}, with open-source implementations accessible to researchers without massive budgets for cloud compute.
Our research provides a \textit{proof of concept} that LLMs can benefit from the addition of epinet, which we expect to transfer to other related tasks.
We've already shown in Figure~\ref{fig:mnli_headline} that epinet can lead to significant improvements on the MNLI dataset;
in fact, these results extend more broadly to the full GLUE suite.

We consider a simple per task fine-tuning setting where the agents are trained on each GLUE task separately.
Our \textbf{baseline} agent follows the procedure from Section~\ref{sec:synthetic}; it is trained by selecting a fixed and random subset of the data.
We sweep over batch size $\in \{4, 16, 64\}$, learning rate $\in \{$1e-6, 3e-6, 1e-5, 3e-5, 1e-4, 3e-4$\}$.
For each setting we perform 10 epochs of SGD training and select the best training step in hindsight.
This baseline agent is not trained online, but is meant to provide an upper bound on how well any agent that does not prioritize its data can do.

%%%%%%%%%%%%%%%%%%%%%%%%%%%%%%%%%%%%%%%%%%%%%%%%%%%%%%%%%%%%%%%%%%% Agents
\subsection{Agents}
\label{sec:language_agent}

% \ian{Need to fill in details about the network architectures we consider.}
% This is not too difficult... essentially we already said it before, but need concrete details on epinet, ensemble, dropout.
% Talk about the BERT initial parameters.
% Essentially it's an MLP that hangs off the final hidden layer of BERT.
% \fillpara
We consider ENNs based around fine-tuning a pretrained BERT architecture.
At a high level, these ENNs work by branching from the final hidden layer in the BERT network.
Table~\ref{tab:agent_summary_bert} outlines these architectures, together with the hyperparameters we tune.

\vspace{-2mm}
\begin{table*}[!ht]
\caption{Summary of agent architectures, full details in Appendix \ref{app:language}.}
\label{tab:agent_summary_bert}
\begin{center}
\resizebox{0.8\textwidth}{!}{%
\begin{tabular}{|l|l|l|}
\hline
% \rowcolor{tableHeader}
\textcolor{black}{\textbf{agent}}   & \textcolor{black}{\textbf{description}}            & \textcolor{black}{\textbf{hyperparameters}} \\[0.5ex]  \hline
\textbf{\texttt{baseline}}          & MLP classification layer \citep{devlin2018bert}        &  learning rate, network                        \\
\textbf{\texttt{dropout}}           & MLP dropout \citep{Gal2016Dropout}                            &  learning rate, network, dropout rate                \\
\textbf{\texttt{ensemble}}          & MLP deep ensembles \citep{lakshminarayanan2017simple}         & learning rate, network, ensemble size                        \\
\textbf{\texttt{epinet}}            & MLP + MLP epinet \citep{osband2021epistemic}              & learning rate, network, prior, index dimension \\ \hline
\end{tabular}%
}
\end{center}
\vspace{-1mm}
\end{table*}

For all of our agents apart from \texttt{epinet}, we found no benefit to using an MLP head over a simple linear layer.
This is perhaps unsurprising since BERT is already a very large model, so one additional hidden layer makes little difference.
For the epinet, we use a network
{
\medmuskip=0mu
\thinmuskip=0mu
\thickmuskip=0mu
\begin{equation}
    \sigma_\eta(\tilde{x}, z) = \left(h_\eta(\mathtt{concat}(\tilde{x}, z)) + \lambda h^P(\mathtt{concat}(\tilde{x}, z)) \right)^T z.
\end{equation}
}
\hspace{-1mm}Here $\tilde{x} := \mathrm{sg}[\phi_\zeta(x)]$ is the last hidden feature for BERT, and $h$ is a 2-layer MLP with 50 hidden units.
The MLP $h^P$ is initialized as a `prior network' and has no trainable parameters \citep{osband2018rpf}, and $\lambda > 0$ is a scaling parameter.
This part of the network serves to drive initial variance in index $z$, but that can be resolved with data.
We push the details of the network implementation and open-source code to Appendix~\ref{app:code}.

We train all of our agents according to Algorithm~\ref{alg:enn_training} by examining $N_B=40$ candidate inputs and then selecting $n_b=4$ examples to train with $n_Z=10$ index samples.
In each case we tune the agent hyperparameters on the MNLI dataset, and then repeat the training and evaluation across the other GLUE tasks.
For all agents, we found that learning rate 1e-5 matched the best performance observed.

%%%%%%%%%%%%%%%%%%%%%%%%%%%%%%%%%%%%%%%%%%%%%%%%%%%%%%%%%%%%%%%%%%% Results
\subsection{Results}
\label{sec:language_results}

\begin{figure}[!ht]
\centering
\subfigure[Epinet outperforms marginal methods.]
{
  \centering
  \includegraphics[width=.95\columnwidth]{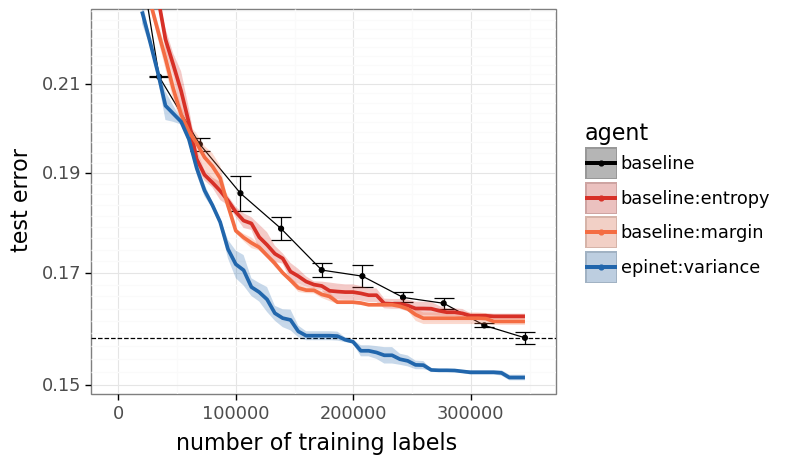}
  \label{fig:mnli_marginal}
  }
\vspace{1mm}
\subfigure[Epinet outperforms other ENNs.]
{
  \centering
  \includegraphics[width=.95\columnwidth]{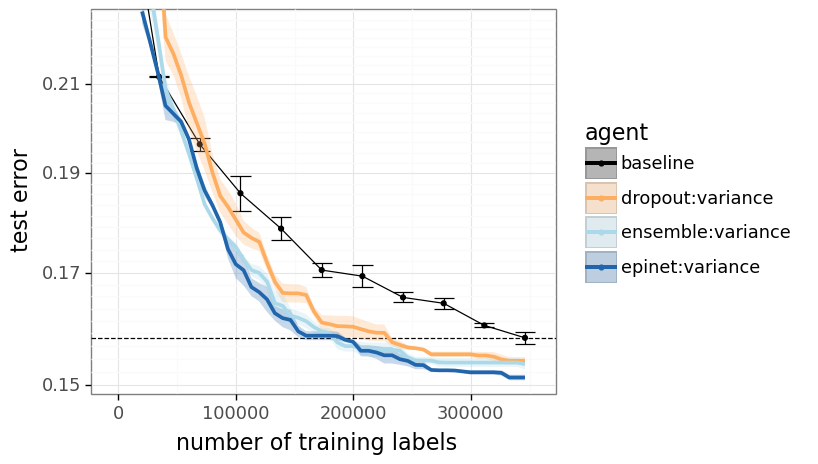}
  \label{fig:mnli_bayes}
  }
\vspace{-5mm}
\caption{Active learning with an epinet is able to provide larger improvements with fewer labels than competing approaches for BERT on MNLI \citep{williams2017broad}.}
\vspace{-1mm}
\label{fig:mnli_comparison}
\end{figure}

% Running through examples on MNL
Figure~\ref{fig:mnli_headline} shows that, using an epinet, we can match baseline performance on the MNLI dataset while using 2x fewer labels.
Figure~\ref{fig:mnli_comparison} shows that these results are impressive even when compared against other benchmark approaches to active learning.
Figure~\ref{fig:mnli_marginal} compares epinet prioritized by variance against other methods that do not make use of epistemic uncertainty.
Their performance plateaus earlier as they prioritize potentially noisy examples, rather than informative training data.
This mirrors our earlier result on the Neural Testbed (Figure~\ref{fig:testbed_marginal}).
Figure~\ref{fig:mnli_bayes} compares agent performance when prioritizing by \textbf{variance}, changing the ENN architecture.
We see that epinet performs better than the competing approaches, and the ensemble performs better than dropout in this setting.
Once again, the qualitative results mirror the synthetic testbed (Figure~\ref{fig:testbed_bayes}).
Unlike the testbed, we see that using an epinet also improves the final performance of the model, not just the required number of training labels to reach that performance.

\vspace{-1mm}

\begin{figure*}[!ht]
    % \vspace{-2mm}
    \centering
    \includegraphics[width=\textwidth]{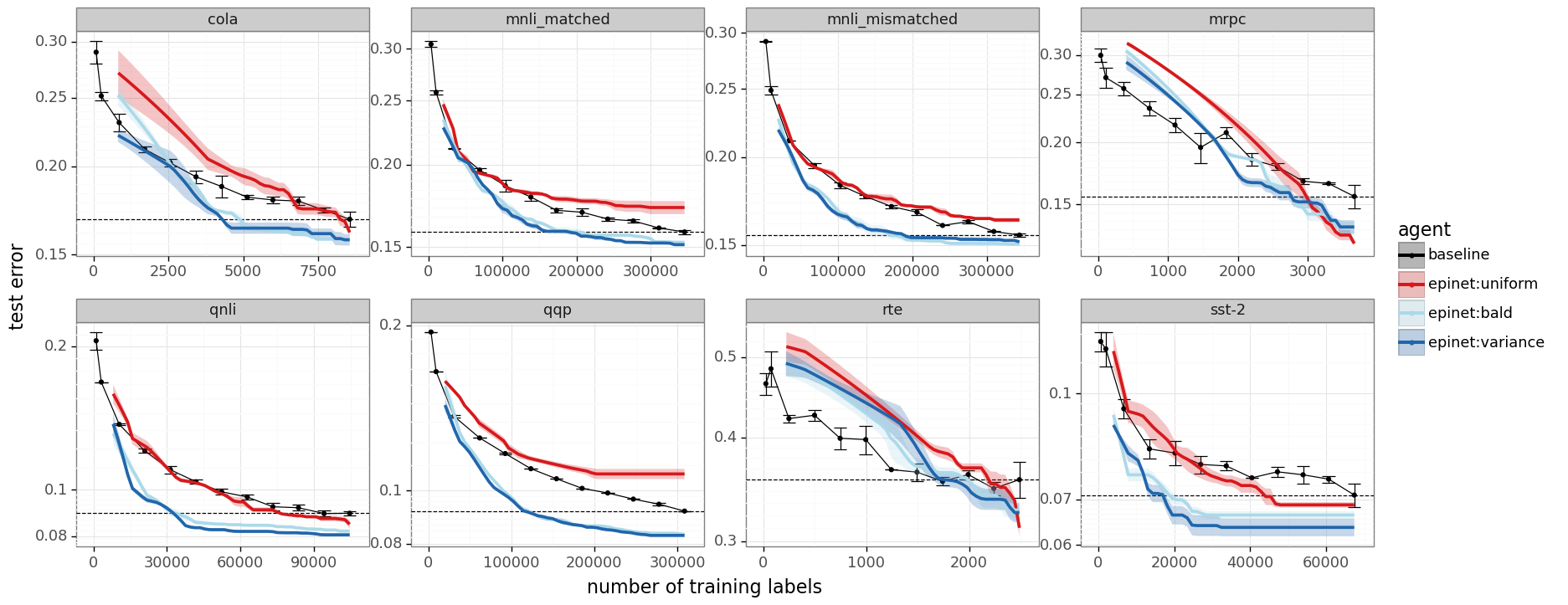}
    \vspace{-7mm}
    \caption{Fine-tuning BERT models across GLUE tasks. The baseline BERT is tuned over learning rate, batch size and SGD steps for each random subset of training data. Using an epinet with epistemic prioritization consistently learns with fewer labels and better final performance. Using an epinet with uniform data selection does not produce the same benefit. The choice of epistemic priority (bald vs variance) is relatively unimportant.}
    \label{fig:epinet_glue}
    \vspace{-1mm}
\end{figure*}

Figure~\ref{fig:epinet_glue} repeats the the analysis of Figure~\ref{fig:mnli_headline} and MNLI on all GLUE classification tasks.
We include results for epinet prioritized by both \textbf{bald} and \textbf{variance}, as well as an epinet with \textbf{uniform} data prioritization.
These bald and variance are essentially indistinguishable from each other in our experiments, and both significantly better than the baseline.
The epinet with uniform data selection performs much worse, showing that the benefits from epinet do not come purely from the network architecture, but also the data selection scheme.
In fact, on many tasks the uniform data selection does not match the baseline final performance.
We provide more details on epinet performance under different priority schemes in Appendix~\ref{app:language}.
% We believe this could be rectified with further learning rate tuning and more epochs of training as per the baseline agent.
% The only two tasks where prioritization by epinet offers relatively little improvement are `mrpc' and `rte'.
% We believe that this is due to our learning rate being tunes for larger datasets, and an artefact of our online training Algorithm~\ref{alg:enn_training} that could be improved if we tuned per-task.

\begin{figure}[!ht]
    % \vspace{-4mm}
    \centering
    \includegraphics[width=0.99\columnwidth]{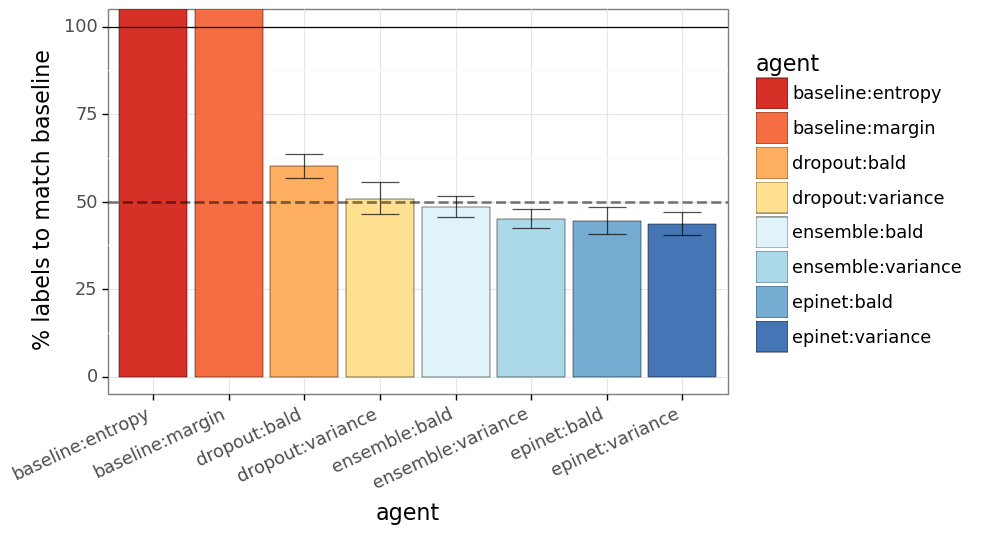}
    \vspace{-7mm}
    \caption{Learning with an epinet matches BERT performance in 2x less data. Using either dropout or ensemble as ENNs also perfoms well, but prioritizing by entropy/margin does not always reach baseline performance.}
    \label{fig:glue_bar}
    \vspace{-4mm}
\end{figure}

Figure~\ref{fig:glue_bar} compares the aggregate performance of several agents over all GLUE classification tasks.
For each task, we train three random seeds according to Algorithm~\ref{alg:enn_training} and compute the number of labels required to match baseline performance on the full dataset.
We then compute the geometric mean of this ratio, together with confidence intervals based on one standard error.
\textbf{On average, learning with an epinet allows us to match the baseline performance with 2x less data than training without prioritization}.
This is true using either \textbf{bald} or \textbf{variance} prioritization.
The heuristic approaches of \textbf{entropy} and \textbf{margin} prioritization do not match the baseline performance on MNLI tasks, and so their ratio is taken to be infinite.
Agents that use either \textbf{dropout} or \textbf{ensemble} ENNs were also able to see significant improvements in data efficiency.
However, both of these network architectures have much higher computational costs than the epinet, and so are less promising for scaling to future LLM research at scale.
We include further analysis and learning curves in Appendix~\ref{app:language}.

%%%%%%%%%%%%%%%%%%%%%%%%%%%%%%%%%%%%%%%%%%%%%%%%%%%%%%%%%%%%%%%%%%%%%%%%%%%%%%%% CONCLUSION
%%%%%%%%%%%%%%%%%%%%%%%%%%%%%%%%%%%%%%%%%%%%%%%%%%%%%%%%%%%%%%%%%%%%%%%%%%%%%%%%
\vspace{-1mm}
\section{Conclusion}
\label{sec:conclusion}

This paper looks at the problem of active learning in fine-tuning language models.
We show that, using an epinet, we can augment LMs to output the kind of uncertainty estimates that are useful to prioritize label acquisition.
On GLUE tasks, this allows us to train a BERT model using 2x fewer labels.
Further, as we continue to train on the full datasets we actually reach better final performance.
These improvements are significant, and at least match those attained by previous approaches to Bayesian deep learning.
Importantly, the epinet architecture offers a way to scale this epistemic uncertainty to large pretrained language models with only model incremental computation.

Although these results are significant, we are motivated by the further potential of LLMs that know what they don't know.
In doing this, we can unlock new families of algorithms and approaches that can may lay the foundation for much larger improvements in data efficiency \citep{lu2021reinforcement}.
We view this paper as an important `proof of concept' that adding epistemic uncertainty estimates to language models is possible in a computationally tractable manner, and that even quite simplistic approaches that leverage this uncertainty can significantly improve data efficiency.

% \newpage
%%%%%%%%%%%%%%%%%%%%%%%%%%%%%%%%%%%%%%%%%%%%%%%%%%%%%%%%%%%%%%%%%%%%%%%%%%%%%%%% ACKNOWLEDGEMENTS
%%%%%%%%%%%%%%%%%%%%%%%%%%%%%%%%%%%%%%%%%%%%%%%%%%%%%%%%%%%%%%%%%%%%%%%%%%%%%%%%
% \section*{Acknowledgements}

% We thank Vikranth Dwaracherla, Vlad Firoiu, and Grace Lam for helpful discussion and feedback in the development of this paper.
% We would also like to thank the entire of the Efficient Agent team at DeepMind for their continued help and development in this project, as well as surrounding infrastructure relating to ENNs and agent development.

%%%%%%%%%%%%%%%%%%%%%%%%%%%%%%%%%%%%%%%%%%%%%%%%%%%%%%%%%%%%%%%%%%%%%%%%%%%%%%%% BIBLIOGRAPHY
%%%%%%%%%%%%%%%%%%%%%%%%%%%%%%%%%%%%%%%%%%%%%%%%%%%%%%%%%%%%%%%%%%%%%%%%%%%%%%%%
\bibliography{references}
\bibliographystyle{icml2023}

%%%%%%%%%%%%%%%%%%%%%%%%%%%%%%%%%%%%%%%%%%%%%%%%%%%%%%%%%%%%%%%%%%%%%%%%%%%%%%%
%%%%%%%%%%%%%%%%%%%%%%%%%%%%%%%%%%%%%%%%%%%%%%%%%%%%%%%%%%%%%%%%%%%%%%%%%%%%%%%
% APPENDIX
%%%%%%%%%%%%%%%%%%%%%%%%%%%%%%%%%%%%%%%%%%%%%%%%%%%%%%%%%%%%%%%%%%%%%%%%%%%%%%%
%%%%%%%%%%%%%%%%%%%%%%%%%%%%%%%%%%%%%%%%%%%%%%%%%%%%%%%%%%%%%%%%%%%%%%%%%%%%%%%
\newpage
\appendix
\onecolumn

%%%%%%%%%%%%%%%%%%%%%%%%%%%%%%%%%%%%%%%%%%%%%%%%%%%%%%%%%%%%%%%%%%%%%%%%%%%%%%%% CODE
%%%%%%%%%%%%%%%%%%%%%%%%%%%%%%%%%%%%%%%%%%%%%%%%%%%%%%%%%%%%%%%%%%%%%%%%%%%%%%%%
\section{Open source code}
\label{app:code}

Two related github repositories complement this paper:
{
\begin{enumerate}[leftmargin=*]
    \item \textbf{\texttt{enn}}: \githubenn
    \item \mbox{\textbf{\texttt{neural\_testbed}}: \githubtestbed}
\end{enumerate}
}
These libraries contain the code necessary to reproduce the key results in our paper, divided into repositories based on focus.
Together with each repository, we include several `tutorial colabs' -- Jupyter notebooks that can be run in a browser without requiring any local installation.
Each of these libraries is written in Python, and relies heavily on JAX for scientific computing \citep{jax2018github}.
We view this open-source effort as a major contribution of our paper.

The first library, \texttt{enn}, was introduced as part of \citet{osband2021epistemic}.
It focuses on the design of epistemic neural networks and their training.
In this submission, we contribute additional code around the BERT network \nolinkurl{enn/networks/bert} as well as the priority functions \nolinkurl{enn/active\_learning}.

The second library, \texttt{neural\_testbed}, was introduced as part of \textit{The Neural Testbed} \citep{osband2022neural}.
We add experiments around the active learning experiments considered in Section~\ref{sec:synthetic}, which can be found in \nolinkurl{neural\_testbed/active\_learning}.

%%%%%%%%%%%%%%%%%%%%%%%%%%%%%%%%%%%%%%%%%%%%%%%%%%%%%%%%%%%%%%%%%%%%%%%%%%%%%%%% SYNTHETIC
%%%%%%%%%%%%%%%%%%%%%%%%%%%%%%%%%%%%%%%%%%%%%%%%%%%%%%%%%%%%%%%%%%%%%%%%%%%%%%%%
\section{Synthetic data}
\label{app:synthetic}

This section provides details about the Neural Testbed experiments in Section~\ref{sec:synthetic}.
We begin with a review of the neural testbed as a benchmark problem, and the associated generative models.
We then give an overview of the baseline agents we compare against in our evaluation.

\subsection{Environment}
The Neural Testbed \citep{osband2022neural} is a collection of neural-network-based, synthetic classification problems that evaluate the quality of an agent's predictive distributions. We make use of the open-source code at \githubtestbedpublic. The Testbed problems use random 2-layer MLPs with width $50$ to generate training and testing data.
The specific version we test our agents on entails binary classification, input dimension $D = 10$, number of training samples $T = 200$, temperature $\rho = 0.1$ for controlling the signal-to-noise ratio, and $10$ random seeds for generating different problems.
The performance metrics are averaged across problems to give the final performance scores.

\subsection{Agents}
We follow \citet{osband2022neural} and consider the benchmark agents as in Table \ref{tab:agent_summary}, which includes the epinet from \citet{osband2021epistemic}. We use the open-source implementation and hyperparameter sweeps at \nolinkurl{/agents/factories}.
According to \citet{osband2022neural}, the benchmark agents are carefully tuned on the Testbed problems, so we do not further tune these agents.

%%%%%%%%%%%%%%%%%%%%%%%%%%%%%%%%%%%%%%%%%%%%%%%%%%%%%%%%%%%%%%%%%%%%%%%%%%%%%%%% LANGUAGE
%%%%%%%%%%%%%%%%%%%%%%%%%%%%%%%%%%%%%%%%%%%%%%%%%%%%%%%%%%%%%%%%%%%%%%%%%%%%%%%%
\section{Language models}
\label{app:language}

This section provides details about the BERT fine-tuning experiments in Section~\ref{sec:language}.
We begin with a review of the `baseline' agent we use for evaluation in BERT.
We then give an overview of hyperparameter tuning and sensitivities in the agents we investigate.

\subsection{Baseline agent}
\label{app:language_baseline}

A key component of our research was to set a strong baseline for BERT in this active learning setting.
Since prior works have focused on a purely supervised learning setting, where the entire training set is made available at once, typical BERT implementations are not optimized for active learning.
To sidestep these issues, we instead use a baseline that \textit{separately} has the opportunity to retrain on different fractions of training data.
For each GLUE task, we sample a fraction $\psi$ of the data, and then optimize a supervised learning BERT baseline trained only on that fraction of the data.
We then obtain the baseline learning curve shown in our figures by sweeping,
$$\psi \in \{0.01, 0.03, 0.1, 0.2, 0.3, 0.4, 0.5, 0.6, 0.7, 0.8, 0.9, 1.\}.$$
For each of these sub-datasets, we sweep over batch size $\in \{4, 16, 64\}$, learning rate $\in \{$1e-6, 3e-6, 1e-5, 3e-5, 1e-4, 3e-4$\}$, and select the best SGD step over 10 training epochs in hindsight.
For each setting, we average the results over three random seeds to obtain standard error estimates for the quality of this baseline.

The performance of this `baseline' agent is therefore much stronger than any naive application of BERT training to active learning.
We allow the supervised learning baseline to optimize its parameters and hyperparameters purely for that problem setting (and data fraction) of interest.
This means that the resultant baseline is really an upper bound on the performance that can reasonably be expected by any learning scheme that does not actively prioritize its training data.

%%%%%%%%%%%%%%%%%%%%%%%%%%%%%%%%%%%%%%%%%%%%%%%%%%%%%%%%%%%%%%%%%%%%%%%%%%%%%%%%%%%%%%%%% ROBUSTNESS
\subsection{Robustness and hyperparameters}

This subsection outlines some of the key hyperparameter tunings and robustness analysis that we performed during our experiments.
Due to computational costs, as well as concerns on overfitting, we tuned our agents on the MNLI matched dataset and then used these settings for the remainder of the GLUE tasks.
We chose MNLI as it was the largest dataset in the GLUE task, and widely regarded as one of the most challenging tasks.
We imagine that it is possible to further improve our results by tuning each approach per task, although this probably is not of much interest in terms of research contribution.

Except where otherwise stated, all of our agents use an ADAM optimizer with learning rate $1e-5$, decay parameter $b_2=0.95$ and clipping global norm of the gradient to be at most 1, apart from that we use the default settings provided by optax \citep{kingma2014adam, deepmind2020jax}.
We train all of our agents according to Algorithm~\ref{alg:enn_training} by examining $N_B=40$ candidate inputs and then selecting $n_b=4$ examples to train on with $n_Z=10$ index samples.
This choice of $N_B=40, N_b=4$ is chosen for computational convenience as we distribute compute across several TPU cores, and running a fully-batched active learning scheme that evaluated much higher $N_B$ and trained on only $N_b=1$ would have taken longer to run in wall clock time.
For each batch we train with $n_Z=10$ index samples.

%%%%%%%%%%%%%%%%%%%%%%%%%%%%%%%%%%%%%%%%%%%%%%%%%%%%%%%%%%%%%%%%%%%%%%%%%%%%% Marginal methods
\subsubsection{Marginal methods}

\begin{figure}[!ht]
\centering
\subfigure[Tuning learning rate for \textbf{entropy}.]
{
  \centering
  \includegraphics[width=.45\linewidth]{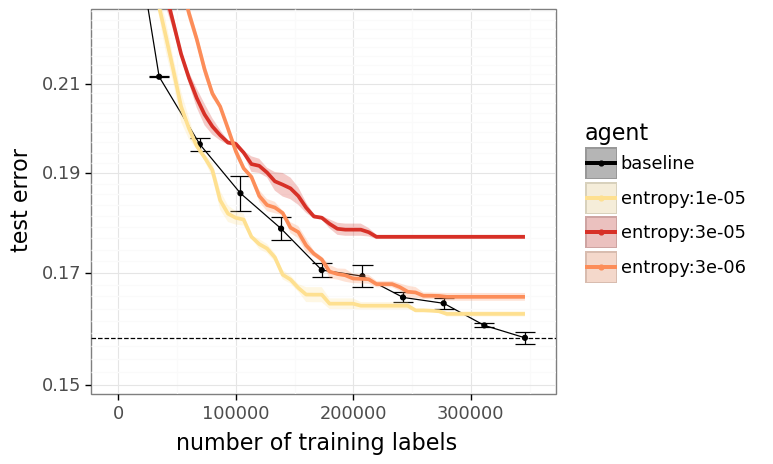}
  \label{fig:entropy_mnli}
  }
\hspace{2mm}
\subfigure[Tuning learning rate for \textbf{margin}.]
{
  \centering
  \includegraphics[width=.45\linewidth]{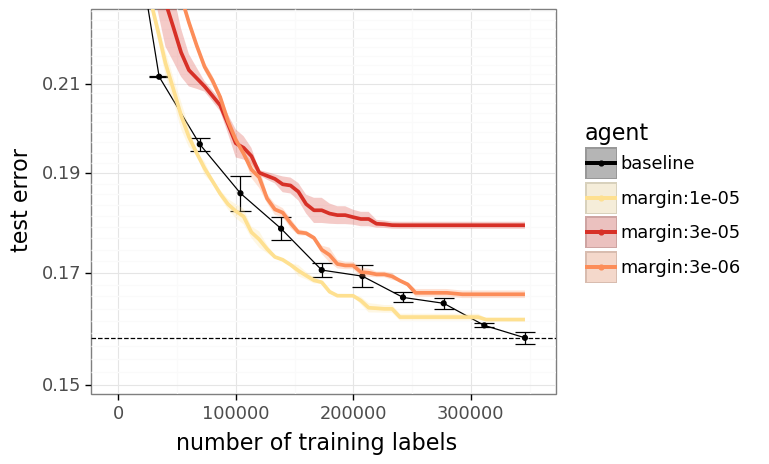}
  \label{fig:margin_mnli}
  }
\vspace{-1mm}
\caption{Tuning learning rate on the MNLI matched dataset. Marginal prioritization methods can sometimes lead to faster learning, but have a tendency to plateau at higher test error. We found that \textbf{margin} and \textbf{entropy} prioritization consistently perform similarly.}
\vspace{-1mm}
\label{fig:marginal_mnli_appendix}
\end{figure}

For our marginal prioritization methods we tuned the learning rate $\in \{$3e-6, 1e-5, 3e-5$\}$.
We found that performance was quite sensitive to this parameter across both \textbf{entropy} and \textbf{margin} priority schemes.
Figure~\ref{fig:marginal_mnli_appendix} compares their performance on the MNLI matched dataset.
We can see that none of the learning rates are able to match the performance of the \textbf{baseline} agent will full data.

%%%%%%%%%%%%%%%%%%%%%%%%%%%%%%%%%%%%%%%%%%%%%%%%%%%%%%%%%%%%%%%%%%%%%%%%%%%%% Dropout
\subsubsection{Dropout}

For the dropout agent we tuned the dropout rate $p \in \{0, 0.05, 0.1, 0.2, 0.5\}$ over three seeds each on the MNLI matched dataset.
We found that the dropout rate 0.1 performed best, and was the only one that matched our baseline performance over all three seeds.

\begin{figure}[!ht]
    % \vspace{-4mm}
    \centering
    \includegraphics[width=0.6\textwidth]{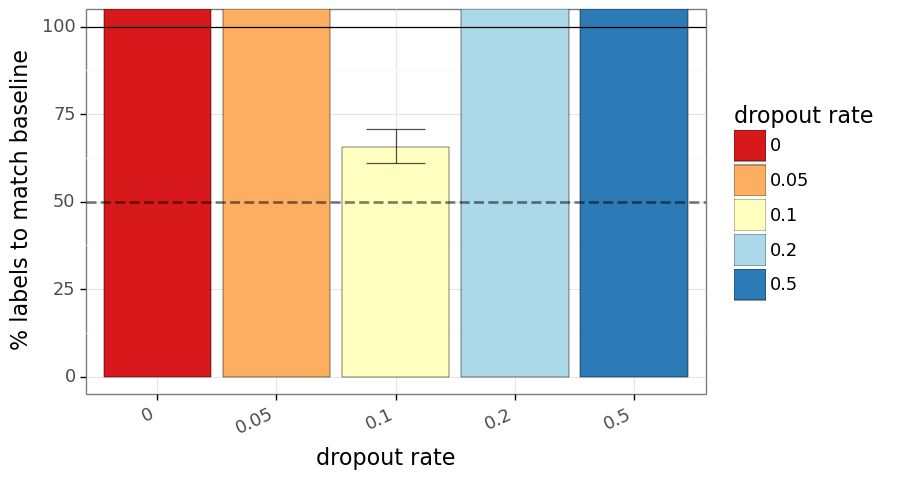}
    \vspace{-2mm}
    \caption{Varying dropout rate on the MNLI matched dataset. Only dropout rate 0.1 is able to match the baseline performance in our experiment.}
    \label{fig:drop_bar}
    \vspace{-2mm}
\end{figure}

% \vspace{-1mm}
% \begin{figure}[!ht]
% \centering
% \begin{subfigure}{.45\textwidth}
%   \centering
%   \includegraphics[width=.99\linewidth]{figures/drop_bar.png}
%   \caption{Data requirements by dropout rate.}
%   \label{fig:drop_bar}
% \end{subfigure}
% \hspace{2mm}
% \begin{subfigure}{.45\textwidth}
%   \centering
%   \includegraphics[width=.99\linewidth]{figures/mnli_drop.png}
%   \caption{Performance of agents through training.}
%   \label{fig:mnli_drop}
% \end{subfigure}
% \vspace{-1mm}
% \caption{Varying dropout rate on MNLI matched dataset.}
% \vspace{-1mm}
% \label{fig:drop_comparison}
% \end{figure}

%%%%%%%%%%%%%%%%%%%%%%%%%%%%%%%%%%%%%%%%%%%%%%%%%%%%%%%%%%%%%%%%%%%%%%%%%%%%% Ensemble
\subsubsection{Ensemble}

For the ensemble agent we vary the ensemble size $K \in \{1, 3, 10, 30\}$ over three seeds each on the MNLI matched dataset.
We found that the ensemble size 10 performed best, in terms of the average number of labels required to match the baseline performance.
In general, we would expect larger ensembles to perform better, but we did run with $n_Z=10$ which may have impacted the results.
We opted for size 10 since we reported our main headline numbers in terms of number of labels required to match the baseline, and wanted to run our experiments quickly in terms of wall clock time.

\begin{figure}[!ht]
    % \vspace{-4mm}
    \centering
    \includegraphics[width=0.6\textwidth]{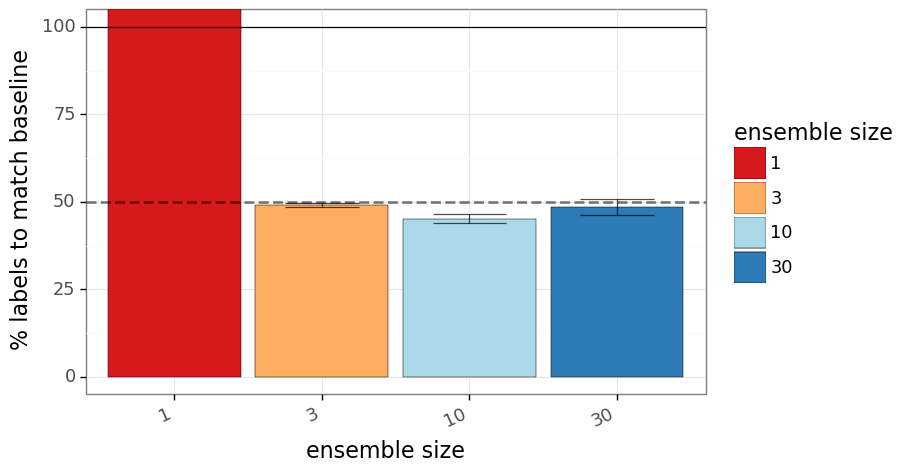}
    \vspace{-2mm}
    \caption{Varying ensemble size on the MNLI matched dataset.}
    \label{fig:ensemble_bar}
    \vspace{-2mm}
\end{figure}

% \vspace{-1mm}
% \begin{figure}[!ht]
% \centering
% \begin{subfigure}{.45\textwidth}
%   \centering
%   \includegraphics[width=.99\linewidth]{figures/ensemble_bar.png}
%   \caption{Data requirements by ensemble size.}
%   \label{fig:ensemble_bar}
% \end{subfigure}
% \hspace{2mm}
% \begin{subfigure}{.45\textwidth}
%   \centering
%   \includegraphics[width=.99\linewidth]{figures/mnli_ensemble.png}
%   \caption{Performance of agents through training.}
%   \label{fig:mnli_ensemble}
% \end{subfigure}
% \vspace{-1mm}
% \caption{Varying ensemble size on MNLI matched dataset.}
% \vspace{-1mm}
% \label{fig:mnli_ensemble}
% \end{figure}

\subsubsection{Epinet}

For the epinet agent we took the `off the shelf' solution provided by the ENN library for ImageNet \citep{osband2021epistemic}.
This consists of a base MLP with 2-layers of 50 hidden units and ReLU activations, and an epinet with the same form and a matched prior function.
Following that work, we used an index dimension of 10 and a prior scale of 1.
Similar to past work, we found that the results were not very sensitive to the choice of these parameters.

\begin{figure}[!ht]
    % \vspace{-4mm}
    \centering
    \includegraphics[width=0.6\textwidth]{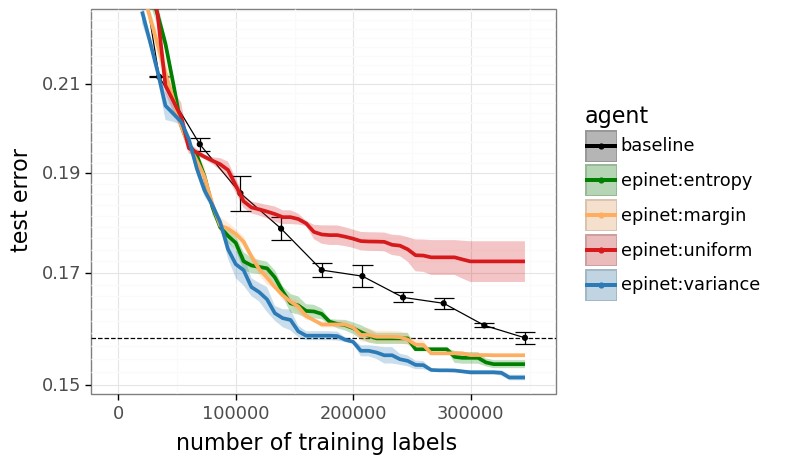}
    \vspace{-2mm}
    \caption{Varying the prioritization scheme on the MNLI matched dataset. Using an epinet, we are able to beat baseline performance using marginal priority methods. However, we can obtain even better performance by focusing on datapoints with high epistemic uncertainty.}
    \label{fig:mnli_epi}
    \vspace{-2mm}
\end{figure}

To investigate the sensitivity of the epinet to the prioritization scheme, we ran the same epinet architecture with each of the prioritization schemes for three seeds on the MNLI matched dataset.
We present these results in Figure~\ref{fig:mnli_epi}.
We can see that the epinet performs better when prioritizing by epistemic uncertainty (variance), although the marginal prioritization methods (margin, entropy) also lead to significant improvements over the baseline.
However, learning with uniform prioritization does not match baseline performance after one epoch.

\begin{figure}[!ht]
    % \vspace{-4mm}
    \centering
    \includegraphics[width=0.6\textwidth]{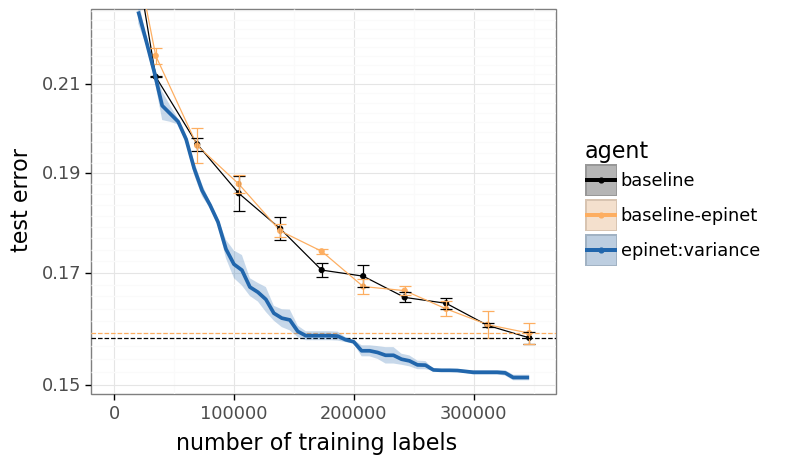}
    \vspace{-2mm}
    \caption{Comparing epinet performance against supervised baselines on the MNLI matched dataset. When trained using supervised learning alone, the baseline-epinet offers no significant benefit in test accuracy over baseline.}
    \label{fig:supervised_epinet}
    \vspace{-2mm}
\end{figure}

The results of Figure~\ref{fig:mnli_epi} are interesting in that they show that the epinet with marginal prioritization performs better than the baseline BERT model with the same priority scheme (Figure~\ref{fig:marginal_mnli_appendix}).
This shows that some of the benefits of \textbf{epinet:variance} come from the epinet architecture \textit{without} prioritizing on epistemic uncertainty.
To get an idea of the benefits purely from the epinet architecture, we re-run the \textbf{baseline} agent of Appendix~\ref{app:language_baseline} but using the epinet architecture.
We call the resultant supervised learning agent the \textbf{baseline-epinet} and compare its performance against the \textbf{baseline} and \textbf{epinet:variance} in Figure~\ref{fig:supervised_epinet}.
Here we can see that the \textbf{baseline-epinet} is essentially indistinguishable from the \textbf{baseline} agent.
This shows that the principle benefits of learning with epinet come from the ability to prioritize data more effectively, rather than purely from improvements in network architecture in supervised learning.

At first glance, these results might appear confusing: how can the epinet lead to better performance with marginal prioritization methods (Figure~\ref{fig:mnli_epi}) when the purely-supervised learning schemes lead to no improvements (Figure~\ref{fig:supervised_epinet})?
Actually, since the MNLI task is a fixed dataset dataset without noisy labels, there is in practice a high correlation between marginal priority and epistemic uncertainty.
As such, even the marginal priority functions are able to benefit from the uncertainty estimates provided by the epinet to drive faster learning in this setting.

%%%%%%%%%%%%%%%%%%%%%%%%%%%%%%%%%%%%%%%%%%%%%%%%%%%%%%%%%%%%%%%%%%%%%%%%%%%%%%%%%%%%%%%%% RESULTS BY TASK
\subsection{Results by task}

\begin{figure}[!ht]
    % \vspace{-4mm}
    \centering
    \includegraphics[width=0.9\textwidth]{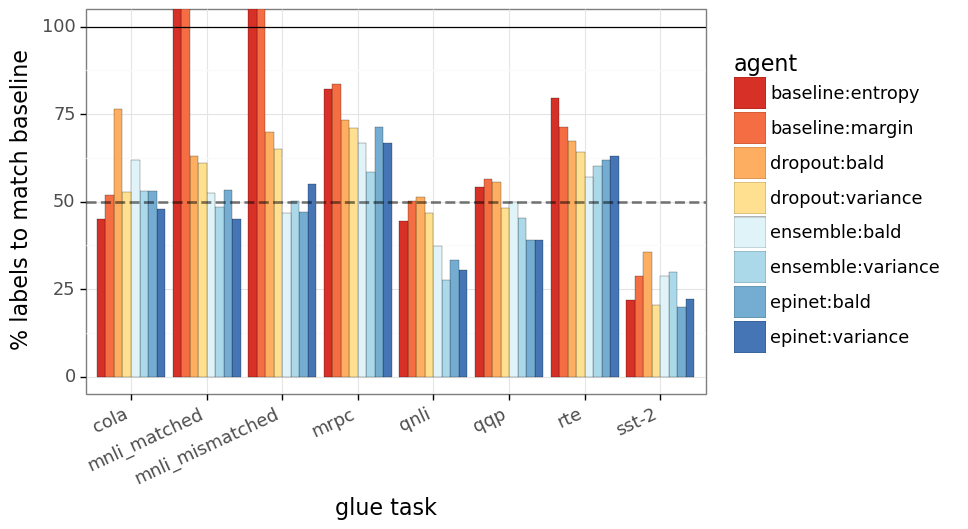}
    \vspace{-2mm}
    \caption{Comparison of amount of labels required to match baseline performance across GLUE tasks. The geometric mean of these results is presented in Figure~\ref{fig:glue_bar}.}
    \label{fig:glue_bar_task}
    \vspace{-2mm}
\end{figure}

Figure~\ref{fig:glue_bar} presents a summary of agent performance across all GLUE tasks.
Figure~\ref{fig:glue_bar_task} breaks down the performance for each agent across each task.
We see that, although there is some variability across tasks, the general pattern of the results is somewhat consistent.
Importantly, marginal priority methods are unable to match the baseline performance even after tuning on MNLI matched or mismatched datasets.
However, on other datasets, they can be tuned to provide reasonable improvements in data-efficiency over the baseline.

\end{document}